# JSMNet: Improving Indoor Point Cloud Semantic and Instance Segmentation through Self-Attention and Multiscale Fusion

Shuochen Xu [a, b], Zhenxin Zhang [a, b,*]

[a] Key Laboratory of 3D Information Acquisition and Application, MOE, Capital Normal University, Beijing 100048, China;
[b] College of Resource Environment and Tourism, Capital Normal University, Beijing 100048, China;

**ABSTRACT**

The semantic understanding of indoor 3D point cloud data is crucial for a range of subsequent applications, including indoor service robots, navigation systems, and digital twin engineering. Global features are crucial for achieving high-quality semantic and instance segmentation of indoor point clouds, as they provide essential long-range context information. To this end, we propose JSMNet, which combines a multi-layer network with a global feature self-attention module to jointly segment three-dimensional point cloud semantics and instances. To better express the characteristics of indoor targets, we have designed a multi-resolution feature adaptive fusion module that takes into account the differences in point cloud density caused by varying scanner distances from the target. Additionally, we propose a framework for joint semantic and instance segmentation by integrating semantic and instance features to achieve superior results. We conduct experiments on S3DIS, which is a large three-dimensional indoor point cloud dataset. Our proposed method is compared against other methods, and the results show that it outperforms existing methods in semantic and instance segmentation and provides better results in target local area segmentation. Specifically, our proposed method outperforms PointNet (Qi et al., 2017a) by 16.0% and 26.3% in terms of semantic segmentation mIoU in S3DIS (Area 5) and instance segmentation mPre, respectively. Additionally, it surpasses ASIS (Wang et al., 2019) by 6.0% and 4.6%, respectively, as well as JSPNet (Chen et al., 2022) by a margin of 3.3% for semantic segmentation mIoU and a slight improvement of 0.3% for instance segmentation mPre.

## 1. Introduction

Three-dimensional indoor point cloud scene understanding is a crucial field for indoor navigation (Diaz-Vilarino et al., 2016; Kim et al., 2018), synchronous positioning and indoor scene modeling (Poux et al., 2018). Semantic and instance joint segmentation of indoor point clouds (Long et al., 2015; Kirillov et al., 2019) is a critical technology required to facilitate these applications.

Indoor point cloud semantic segmentation assigns labels to the different areas in the scene based on predefined category labels. On the other hand, instance segmentation is an extension of semantic segmentation, and it allows for further differentiation within a category by labeling different instances. In recent years, there has been rapid development in point cloud labeling methods that leverage deep learning (Wang et al., 2018; Yi et al., 2019). With advancements in three-dimensional information collection technology, indoor point clouds can provide rich geometry, shape, and texture information, thereby presenting realistic scenes. However, the disorder and unstructured nature of 3D point cloud data present significant challenges in efficiently expressing indoor point cloud data semantically. Challenges faced include difficulties in extracting point cloud features and high computational and memory consumption in models.

* Corresponding author.
  E-mail address: zhangzhx@cnu.edu.cn (Z. Zhang).

The application of self-attention mechanism in point cloud understanding and recognition (Xie et al., 2018; Feng et al., 2020) has enabled the efficient extraction of point cloud features. By incorporating self-attention mechanism, semantically rich features can be obtained, which helps to improve the performance of semantic segmentation.

In this study, we present a novel deep learning network framework for accomplishing joint semantics and instance segmentation of indoor point clouds. Our proposed framework combines Transformer and PointConv to create a global feature self-attention coding module as a means of feature extraction, resulting in robust indoor point cloud feature expression. To overcome the loss after information interpolation, we integrate information with different resolutions and adaptively fuse multi-resolution features of each point to increase feature significance. Finally, semantic and instance segmentation modules are integrated into a unified model, allowing the two branches to promote each other, resulting in a better semantic expression effect for indoor point clouds. The following are the contributions of our study:

• We combine Transformer and PointConv to design a global feature encoding layer that is based on self-attention and create an innovative model for point cloud feature extraction.

• We design a multi-resolution feature adaptive fusion module that is specialized for indoor point clouds so that fine, multi-scale, significant feature expression effect can be obtained.

• We propose a new deep learning framework for joint instance and semantic segmentation in which case segmentation and semantic segmentation promote each other.

• Our framework achieves state-of-the-art results on 3D instance segmentation and semantic segmentation tasks on the Stanford large 3D indoor space dataset（S3DIS）(Armeni et al., 2016).

## 2. Method

This study's overall structure is depicted in Figure 1. We use indoor point clouds as input to obtain the instance and semantic labels of each point. To achieve this, we introduce an innovative combination of global feature self-attention module with a multi-layer network. Through the combination of point cloud semantics and instance tasks, we construct a network model with three modules: the Transformer encoding/decoding module, the multi-resolution feature fusion module, and the joint segmentation (semantic and instance) module. In the encoding part of the Transformer encoder-decoder module, we include a set abstraction (SA) layer and three encoding layers of PointConv with an embedded Transformer module behind each layer. This module then outputs the feature matrix via two parallel decoders. Afterward, the multi-resolution feature adaptive fusion module outputs a feature matrix. We use three branches to integrate and supplement information from the two paths, resulting in a better indoor scene segmentation effect.

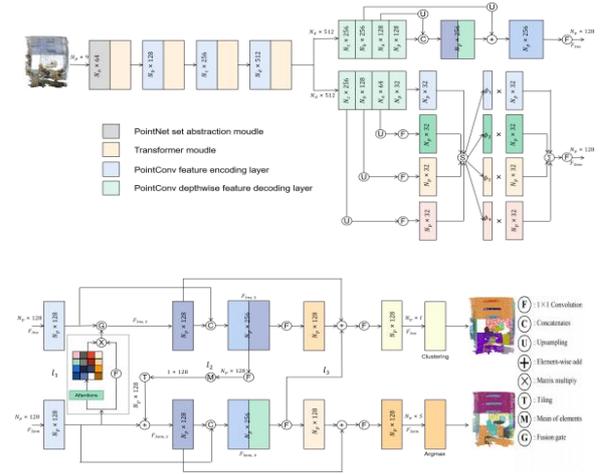

**Figure 1**. An overview of JSMNet, which utilizes self-attention and multiscale fusion for joint semantic and instance segmentation of indoor point clouds.

### 2.1. Transformer encoder-decoder module

In this section, we propose a Transformer encoder-decoder module for initializing features of indoor point clouds. The module consists of two parts: the Transformer encoder and decoder modules. The coding module is constructed by successively applying three encoding layers of PointConv (Wu et al., 2019) after the SA layer (Qi et al., 2017b), followed by four Transformer modules.

### 2.1.1. SA layer

To achieve an abstract representation of point set in indoor point clouds, we introduce the SA layer of PointNet++ (Qi et al., 2017b) in the first layer of the encoder. To achieve an abstract representation of point set in indoor point clouds, we introduce the SA layer of PointNet++ (Qi et al., 2017b) in the first layer of the encoder. Also to avoid causing some loss of useful information,, unlike pointnet++, we use attention pooling (Yang et al.,2020; Hu et al.,2020) instead of maximum pooling to aggregate useful information by automatic learning.

### 2.1.2. Transformer module

The irregular embedding of indoor point clouds in metric space, as well as their insensitivity to the arrangement and cardinality of input features, make Transformer a well-suited architecture for point cloud processing. In our Transformer layer, we We subtract the attention vector $\varphi(x_i)$ from the attention vector $\psi(x_j)$ and add the position encoding $\delta$ to it and to the attention vector $\alpha(x_j)$, given by:

$$y_i = \sum_{x_j \in \mathcal{X}(i)} \rho(\gamma(\varphi(x_i) - \psi(x_j) + \delta)) \odot (\alpha(x_j) + \delta) \qquad (1)$$
$$\delta = \theta(P_i - P_j) \qquad (2)$$

Here, the subset $\mathcal{X}(i) \subseteq \mathcal{X}$ is the local neighborhood of $x_i$ ($k$ nearest neighbors of the $x_i$). The mapping function $\gamma(\cdot)$ is an MLP with two linear layers and a ReLU nonlinear activation function, and $\rho(\cdot)$ is a softmax function. $\varphi(x_i)$, $\psi(x_j)$, and $\alpha(x_j)$ represent three different attention vectors in the transformer. $P_i$ and $P_j$ denote the three-dimensional coordinates of points $i$ and $j$.

We construct a Transformer module centered on a Transformer layer, including a Point Transformer layer, a linear projection, and a residual link to reduce dimensionality and speed up processing.

### 2.2. Multi-resolution feature fusion module

After extracting features using the Transformer codec module, we obtain an output feature matrix with dimensions $N_p \times 512$. The next steps involve upsampling and feature fusion via two separate decoder branches that use the PointConv's depthwise feature decoding layers.

In the semantic branch, we use the same operations as JSNet. Although downsampling the point cloud during multilevel segmentation, fusion, and aggregation of depth information for point cloud tasks such as instance segmentation can benefit the extraction of discriminating features, the corresponding output features could become implicit and abstract. Therefore, we need to recover the feature map that supplies the original points and fully explains the encoded information for each point in the instance branch. To achieve this, we select and fuse fine-grained representations from multi-resolution feature maps. First, the feature matrices of four layers with different resolution dimensions are upsampled to obtain the full-size feature representations of all N points and reconstructed into full-size feature maps by MLP. To further improve our integration method, we analyzed the point-level perceptions $\varphi_m \{\varphi_1, \varphi_2, \varphi_3, \varphi_4\}$ and regressed the fusion parameters $\Phi_m \{\Phi_1, \Phi_2, \Phi_3, \Phi_4\}$ corresponding to the full-sized feature maps $S'_m \{S'_1, S'_2, S'_3, S'_4\}$. Finally, we integrated a comprehensive feature graph $S_{out}$ for instance segmentation into the multi-resolution features of each point:

$$\varphi_m = FC^1(S'_m), m = 1,2,3,4 \qquad (3)$$
$$\phi = softmax(concat(\varphi_1, \varphi_2, \varphi_3, \varphi_4)) \qquad (4)$$
$$S_{out} = \sum_{m=1}^{M} \Phi_m \times S'_m \qquad (5)$$

Where $\forall S'_m \in \{S'_1, S'_2, S'_3, S'_4\}$, FC $(\cdot)$ is a fully-connected layer and its superscript indicates the number of kernels, the point-level information $\Phi_m \in R^N$.

### 2.3. Joint instance and semantic segmentation module

The combination of instance segmentation and semantic segmentation has introduced a unique approach to point cloud instance segmentation. Semantic segmentation and instance segmentation can benefit from each other's learned features. However, previous studies have shown that directly merging instance and semantic information may introduce low-quality semantic information, which can negatively influence the subsequent segmentation tasks (Hou et al., 2019). To address this challenge, we added a branch ($l_1$ in Figure 1) to branches $l_2$ and $l_3$, inspired by recent research called the attentional context fusion module(Wen et al., 2020). In the $l_1$ branch, we employ a self-attention mechanism to blend the original semantic features by weighted average. This attentively enhances the useful information and masks the irrelevant information, thus avoiding the introduction of low-quality semantic information to instance segmentation.

### 2.3.1. Instance branch

As shown in equation (6), In the instance branch, we first pass the feature matrix $F_{sem}$ obtained from the semantic branch through an attention context fusion module and fusion gate, which extracts the features of the semantic branch into the feature matrix of the instance branch to obtain a new feature matrix $F_{ins,\ 2}$. The gating function $Gated$ (·) used in the fusion gate can be found in the research of Wen et al. (2020). Next, As shown in equation (7), we concatenate $F_{ins,\ 2}$ with $F_{INS}$ to obtain the feature matrix $F_{ins,\ 3}$. We then convolve $F_{ins,\ 3}$ using a 1 × 1 convolution and add the resultant to the feature matrix $F_{ins,\ 2}$ and $Conv1D$ (i.e., the operation of $l3$ in Figure 1).

As shown in equation (8), the instance embedding matrix $F'_{ins}$ is obtained by applying a 1 × 1 convolution as described above. Finally, we use the mean-shift clustering algorithm (Comaniciu et al., 2002) to obtain each instance.

$$F_{ins,\ 2} = Gated(F_{ins}, Attention(F_{sem})) \quad (6)$$
$$F_{ins,\ 3} = Concat(F_{ins}, F_{ins,\ 2}) \quad (7)$$
$$F'_{ins} = Conv1D(F_{ins,\ 2} + Conv1D(F_{ins,\ 3}) + Conv1D(F_{sem,\ 3})) \quad (8)$$

### 2.3.2. Semantic branch

In the semantic segmentation branching, we follow a similar approach to the study by Zhao and Tao (2020), by integrating the feature matrix $F_{ins,\ 3}$ generated from the instance branch into the semantic feature space through branching $l_2$. Specifically, we perform a 1 × 1 convolution on $F_{ins,\ 3}$ and then cross-average the resulting matrix column-wise (by taking the mean across elements in each column). Next, as shown in equation (9), we perform a tiling operation (namely the $Tile(\cdot)$) to replicate the resulting matrix row-by-row, generating a matrix that we add element-wise to $F_{sem}$, resulting in matrix $F_{sem,\ 2}$. As shown in equation (10), We obtain the matrix $F_{sem,\ 3}$ by concatenating $F_{sem}$ and $F_{sem,\ 2}$. Finally, similar to the instance branch, as shown in equation (11), we obtain the semantic feature matrix $F'_{sem}$, and use a learned classifier to derive the final semantic labels.

$$F_{sem,2} = F_{sem} + Tile(Mean(Conv1D(F_{ins,3}))) \quad (9)$$
$$F_{sem,\ 3} = Concat(F_{sem}, F_{sem,\ 2}) \quad (10)$$
$$F'_{sem} = Conv1D(F_{sem,2} + Conv1D(F_{sem,2})) \quad (11)$$

## 2.4. Training

During training, the semantic segmentation branch uses cross entropy loss function ($L_{Sem}$). In the instance segmentation branch, as shown in equation (12), we define the loss function as the sum of three parts: $L_{near}$, $L_{far}$, and $L_{reg}$:

$$L_{Ins} = L_{near} + L_{far} + \gamma L_{reg} \quad (12)$$

where $L_{near}$ pulls the embedding item toward the average embedding of the instance (i.e., its center), promoting intra-instance similarity and discouraging inter-instance confusion. $L_{far}$ makes the embeddings of instances exclusive, ensuring that different instances are well-separated and preventing overlap. $L_{reg}$ is a regularization term that maintains the boundary of the embedding values by penalizing embeddings that are too far from the origin, ensuring that the centers of each cluster in the mapping space are not too far away. In the experiments, we set $\gamma$ to 0.001. The formulas are as follows:

$$L_{near} = \frac{1}{I}\sum_{i=1}^{I}\frac{1}{N_i}\sum_{j=1}^{N_i}[\|\tau_i - k_j\|_1 - \delta_v]_+^2 \quad (13)$$

$$L_{far} = \frac{1}{I(I-1)}\sum_{i=1}^{I}\sum_{j=1}^{I}[2\delta_d - \|\tau_i - \tau_j\|_1]_+^2 \quad (14)$$

$$L_{reg} = \frac{1}{I}\sum_{i=1}^{I}\|\tau_i\|_1 \quad (15)$$

$$\tau_i = \frac{1}{N}\sum_{j=1}^{N_i}k_j \quad (16)$$

In the above formulas, the "$I$" represents the number of ground-truth instances, $N_i$ is the number of points in the $i$-th instance, and $k_j$ is the embedding of the $j$-th point. $\tau_i$ represents the average embedding of the instance $i$, which serves as the instance center. $\delta_v$ and $\delta_d$ represent the margins of $L_{near}$ and $L_{far}$, respectively. The notation $[\widehat{A\cdot}]_+ = max\ (0,$

$A$) refers to the hinge function, while $\|\widehat{A}\cdot\|_1$ is the $L_1$ distance. During training, the total loss function is composed of both the semantic branch loss ($L_{Sem}$) and the instance branch loss ($L_{Ins}$), yielding $L = L_{Sem} + L_{Ins}$.

## 3. Experiments

### 3.1. Datasets and evaluation metrics

We assessed the results of our experiments using S3DIS (Stanford large-scale 3D Interior space) dataset (Armeni et al., 2016). This dataset comprises a collection of 3D point clouds representing interior spaces, where each point is defined by its spatial coordinates, spectral information (e.g., RGB values), and semantic/instance labels. The dataset consists of six zones (1-6), encompassing a total of 272 rooms and featuring 13 different item categories.

For the S3DIS dataset, we conducted a 6-fold cross-validation (6-fold CV) based on k-fold cross-validation in PointNet (Qi et al., 2017a) to ensure fair comparison with other methods. Additionally, we present the results of a fifth fold (Area 5), which is a separate building and exhibits some differences from other areas, similar to Tchapmi et al.'s study (2017). For semantic segmentation evaluation, we report overall accuracy (*oAcc*), average accuracy (*mAcc*), and average IoU (*mIoU*). In terms of indoor instance segmentation, indicators include mean precision (*mPrec*), mean recall (*mRec*), and mean coverage with an IoU threshold of 0.5. We also employ the coverage (*Cov*) and weighted coverage (*WCov*) metrics proposed by Ren and Zemel (2017) to assess the performance of indoor scene instance segmentation.

### 3.2. Evaluation and comparison

In this section, we comprehensively evaluate our method and compare it with some existing semantic and instance segmentation methods.

#### 3.2.1. Quantitative results on the S3DIS dataset

**Semantic segmentation.** Our method demonstrates superior semantic segmentation capability on the S3DIS dataset, as shown in Table 1. Compared to other advanced techniques, such as PointNet (Qi et al., 2017a), ASIS (Wang et al., 2019b), JSNet (Zhao and Tao, 2020), and JSPNet (Chen et al., 2022), our approach yields significant advantages. Specifically, our method outperforms JSPNet by 3.1%, 1.3%, and 3.3% in mAcc, oAcc, and mIoU indices in Area 5. In 6-fold cross-validation, our network achieves a 1.7%, 1.5%, and 1.7% improvement in mAcc, oAcc, and mIoU over JSPNet. Moreover, our method almost achieves better overall performance than the latest method.

|  | Method | mAcc(%) | oAcc(%) | mIoU(%) |
|---|---|---|---|---|
| Area 5 | PointNet | 52.1 | 83.5 | 43.4 |
|  | ASIS | 60.9 | 86.9 | 53.4 |
|  | JSNet | 61.4 | 87.7 | 54.5 |
|  | JSPNet | 63.8 | 88.2 | 56.1 |
|  | Ours | 66.9 | **89.5** | **59.4** |
| 6-flod CV | PointNet | 60.3 | 80.3 | 48.9 |
|  | ASIS | 70.1 | 86.2 | 59.3 |
|  | JSNet | 71.7 | 88.7 | 61.7 |
|  | JSPNet | 72.6 | 89.7 | 62.5 |
|  | Ours | **74.3** | **91.2** | **64.2** |

**Table 1.** Semantic segmentation results on S3DIS dataset.

**Instance segmentation.** We evaluated the performance of our method on instance segmentation, and Table 2 presents a comparison with state-of-the-art approaches on the S3DIS dataset. Our method showed improvements in mConv, mWconv, mRecall, and mPre of JSPNet by 0.7%, 0.7%, 0.2%, and 0.3% in Area5. Additionally, in 6-fold cross-validation, our method demonstrated certain advantages. The superior segmentation effect of our approach is visually apparent in Figure 2. For instance, the table and wall segmentations produced by JSNet were confusing, where as our method achieved a better segmentation effect. This emphasizes the superiority of our method.

|  | Method | mCov(%) | mWcov(%) | mRec(%) | mPre(%) |
|---|---|---|---|---|---|
| Aea5 | SGPN | 32.7 | 35.5 | 28.7 | 36.0 |
|  | ASIS | 44.6 | 47.8 | 42.4 | 55.3 |
|  | JSNet | 48.7 | 51.5 | 46.9 | **62.1** |
|  | JSPNet | 50.7 | 53.5 | 48.0 | 59.6 |
|  | Ours | **51.4** | **54.2** | **48.2** | 59.9 |
| 6-fold CV | SGPN | 37.9 | 40.8 | 31.2 | 38.2 |
|  | ASIS | 51.2 | 55.1 | 47.5 | 63.6 |
|  | JSNet | 54.1 | 58.0 | 53.9 | **66.9** |
|  | JSPNet | 54.9 | 58.8 | 55.0 | 66.5 |
|  | Ours | **55.2** | **59.8** | **55.6** | 66.8 |

**Table 2.** Instance segmentation results on S3DIS dataset.

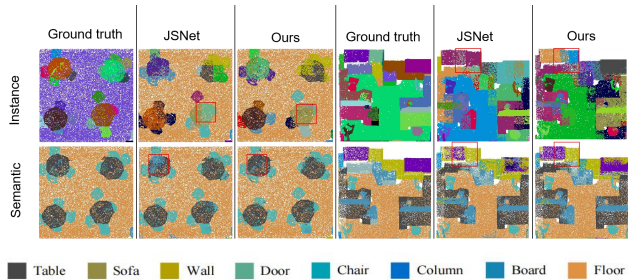

**Figure 2.** Comparison of segmentation results between our method and JSNet on S3DIS.

### 4. Conclusions

This study presents a novel deep learning framework for joint semantics and instance segmentation of indoor point clouds. The framework comprises the Transformer module, multi-resolution feature fusion module, and feature channel aggregation module, specifically designed to enable the joint processing of semantic segmentation and instance segmentation of indoor scenes. By integrating and promoting each other, the two branches of semantic segmentation and instance segmentation achieve superior performance compared to other methods, as revealed by the results of testing on the S3DIS dataset. Our method outperformed JSPNet by 3.1%, 1.3%, and 3.3% on mAcc, oAcc, and mIoU indexes, respectively. Moreover, our method achieved 0.7%, 0.7%, 0.2%, and 0.3% better performance than JSPNet's mConv, mWconv, mRecall, and mPre metrics.

However, indoor scene segmentation remains a challenging task due to high occlusion, clutter, and variability. To obtain more effective and robust results, we plan to consider integrating the multi-level and multi-scale information structure (Tao et al., 2020) and leveraging some fine-grained perceptual models in future work.